\RequirePackage[2020-02-02]{latexrelease}
\documentclass{clv3}
\usepackage{hyperref}
\issue{1}{1}{2020}

\runningtitle{Confusion Probes}

\runningauthor{Shen and Kejriwal}

\begin{document}

\title{Understanding Prior Bias and Choice Paralysis in Transformer-based Language Representation Models through Four Experimental Probes}

\author{Ke Shen\thanks{E-mail: keshen@isi.edu.}}
\affil{Information Sciences Institute, University of Southern California, 4676 Admiralty Way, Suite 1001, Marina del Rey, California 90292}

\author{Mayank Kejriwal\thanks{Corresponding Author: kejriwal@isi.edu}}
\affil{Information Sciences Institute, University of Southern California, 4676 Admiralty Way, Suite 1001, Marina del Rey, California 90292}

%
%
\maketitle

\begin{abstract}
Recent work on transformer-based neural networks has led to impressive advances on multiple-choice natural language understanding (NLU) problems, such as Question Answering (QA) and abductive reasoning. Despite these advances, there is limited work still on understanding whether these models respond to perturbed multiple-choice instances in a sufficiently robust manner that would allow them to be trusted in real-world situations. We present four confusion probes, inspired by similar phenomena first identified in the behavioral science community, to test for problems such as prior bias and choice paralysis. Experimentally, we probe a widely used transformer-based multiple-choice NLU system using four established benchmark datasets. Here we show that the model exhibits significant prior bias and to a lesser, but still highly significant degree, choice paralysis, in addition to other problems. Our results suggest that stronger testing protocols and additional benchmarks may be necessary before the language models are used in front-facing systems or decision making with real world consequences.  
\end{abstract}

\section{Background}

Question Answering (QA) \cite{hirschman2001natural} and inference are important problems in natural language processing (NLP) and applied AI, including development of conversational `chatbot' agents \cite{siblini2019multilingual}. Developments over the last five years in deep neural transformer-based models have led to significant improvements in QA performance, especially in the multiple-choice setting. Bidirectional Encoder Representations from Transformers (BERT) \cite{bert} is a neural transformer-based model that was pre-trained by Google and that consequently achieved state-of-the-art performance in a range of NLP tasks, including QA and Web search. BERT is designed to help computers understand the meaning of ambiguous language in the text by using the surrounding text to establish context, and depends on capabilities such as bidirectional encoding capability, masked language modeling (MLM) and next sentence prediction. 

BERT, and other models based on BERT, such as Patentbert \cite{patentbert}, Docbert \cite{docbert}, SciBERT \cite{scibert}, DistilBERT \cite{distilbert} and K-bert \cite{kbert}, have achieved groundbreaking results in diverse language understanding tasks, including QA \cite{qa1, qa2, qa3}, text summarization \cite{ts1, ts2}, sentence prediction \cite{sp1, sp2}, dialogue response generation \cite{drg1, drg2}, natural language inference \cite{nli1, nli2}, and sentiment classification \cite{sc1, sc2, sc3}. The model studied in this paper, RoBERTa \cite{liu2019roberta}, is a highly optimized version of the original BERT architecture that was first published in 2019 and improved over BERT on various benchmarks by margins of 0.9 [on the Quora Question Pairs dataset \cite{qqp}] - 16.2 percent [on the Recognizing Textual Entailment dataset \cite{dagan2005pascal, haim2006second, giampiccolo2007third, bentivogli2009fifth}]. 

Specifically, RoBERTa is trained with larger mini-batches and learning rates, removes the next-sentence pre-training objective, and focuses on improving the MLM objective to deliver improved performance, compared to BERT, on problems such as Multi-Genre Natural Language Inference \cite{williams2017broad}, and Question-Based Natural Language Inference \cite{rajpurkar2016squad}. RoBERTa-based models have approached near-human performance on various (subsequently described) commonsense natural language understanding (NLU) benchmarks. 

BERT's original success on these NLU tasks has also motivated researchers to adapt it for multi-modal language representation \cite{lu2019vilbert, sun2019videobert}, cross-lingual language models \cite{lample2019cross}, and domain-specific language models, including in the medicine- \cite{alsentzer2019publicly, wang2020cord} and biology-related  domains \cite{biobert}. 
Due to this widespread use, and the fact that even recent, more advanced models based on billions of parameters are based on similar technology (deep transformers), it has become important to systematically study the linguistic properties of BERT using a battery of tests inspired by work first conducted in the behavioral sciences. In prior work, for example, several proposed approaches aimed to study the knowledge encoded within BERT, including fill-in-the-gap probes of MLM \cite{rogers2020primer, wu2019mask}, analysis of self-attention weights \cite{kobayashi2020attention, ettinger2020bert}, the probing of classifiers with different BERT representations as inputs \cite{liu2019linguistic, warstadt2020can}, and a `CheckList' style approach to systematically evaluate the linguistic capability of a BERT-based model \cite{ribeiro2020beyond}. Evidence from this line of research suggests that BERT encodes a hierarchy of linguistic information, with surface features at the bottom, syntactic features in the middle and semantic features at the top \cite{jawahar2019does}. It `naturally' learns syntactic information from  pre-training text. 

However, it has been found that while information can be recovered from its token representation \cite{wu2020perturbed}, it does not fully `understand' naturalistic concepts like negation, and is insensitive to malformed input \cite{rogers2020primer}. The latter is similar to \emph{adversarial} experiments (not dissimilar to adversarial experiments in the computer vision community) that researchers have conducted to test BERT's robustness. Some of these experiments have shown that, even though BERT encodes information about entity types, relations, semantic roles, and proto-roles well, it struggles with the representations of numbers \cite{wallace2019nlp} and is also brittle to named entity replacements \cite{balasubramanian2020s}. Besides, \cite{shen2021generalization} also found clear evidence that fine-tuned BERT-based language representation models still do not generalize well, and may, in fact, be susceptible to dataset bias.

We describe a novel set of systematic \emph{confusion probes} to test linguistically relevant properties of a standard, and currently widely-used, multiple-choice NLU system based on RoBERTa\footnote{Further detailed, with links to the publicly available code, in \emph{Methods}.}. Our probes, described below, are not only replicable, but can be extended to other benchmarks and even newer language representation models as we show through preliminary additional experiments (described in \emph{Discussion}) involving the recent T5-11B model. Unlike much of the prior work on this subject, we are not seeking to understand the layers of a specific network or how it encodes knowledge, but rather, to understand the \emph{commonsense} properties of these models. A clear understanding of such properties allows us to test whether such language models, which are continuing to be rolled out in commercial products, are truly answering questions in a robust manner, or are disproportionately impacted by problems such as \emph{prior bias} and \emph{choice paralysis}. We subsequently define these notions more precisely, but intuitively, \emph{prior bias} occurs when a language representation model has a \emph{consistent and statistically significant} preference for selecting an incorrect candidate choice over another. Such a bias is usually undesirable, as it indicates the model may be amenable to being `tricked' e.g., by introducing perturbations of the kind explored in this paper. 

\emph{Choice paralysis}, on the other hand, occurs when the preference of the model for the \emph{correct} candidate choice significantly and consistently diminishes as more (incorrect) choices are offered to a model in response to a prompt. Choice paralysis is inspired by a similarly named phenomenon in the behavioral and decision sciences\footnote{Other common names include \emph{overchoice} and \emph{choice overload}.} A related (although somewhat broader) problem in decision sciences is \emph{analysis paralysis} \cite{lenz1985}, wherein it was found that giving people too many options can make it more difficult for them to choose between them. We experimentally test whether an analogous problem is observed in multiple-choice NLI QA systems \cite{schwartz2004paradox, TFS}.

\begin{figure}[ht]
\centering
\includegraphics[width=10cm]{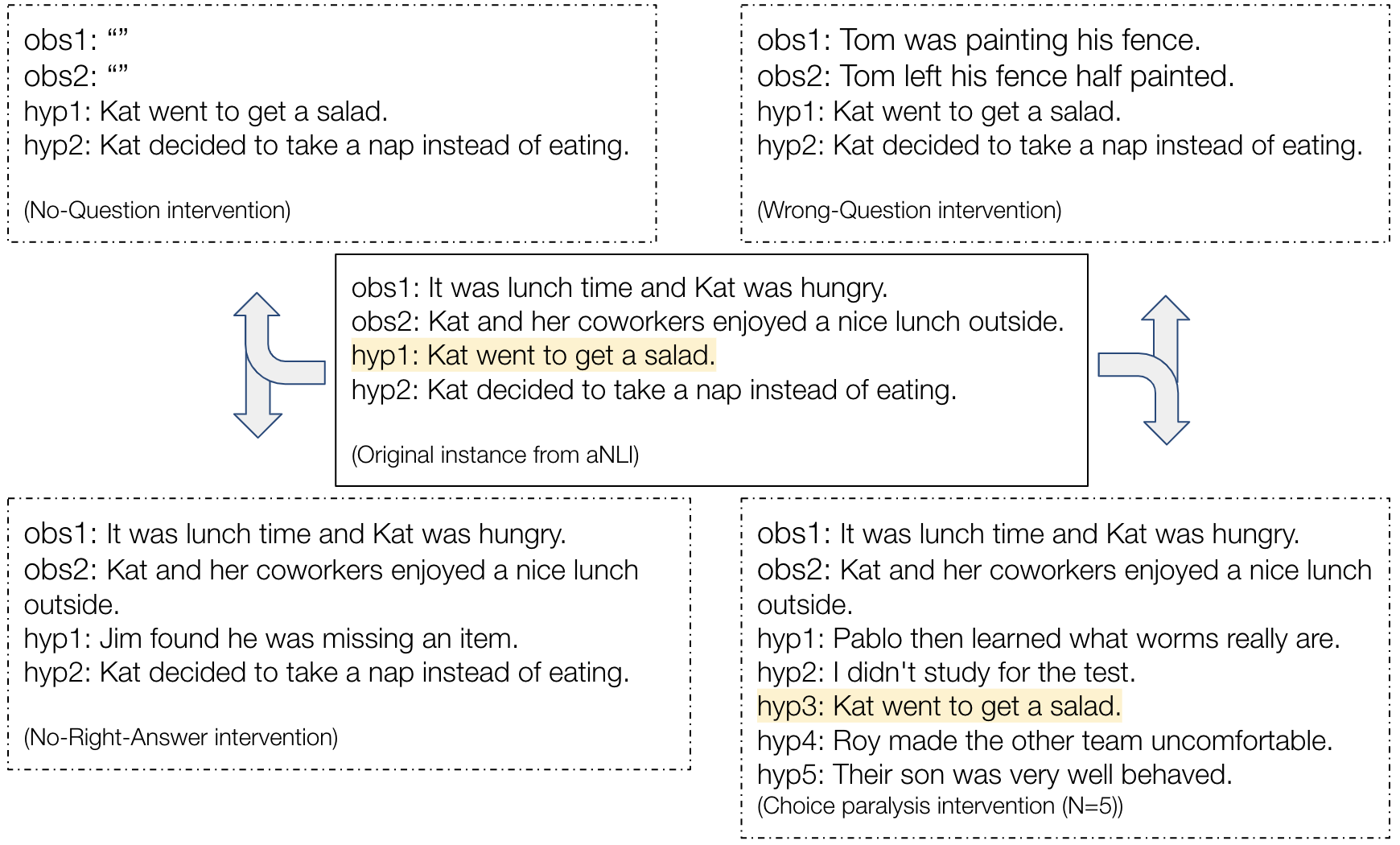}
\caption{An example (from the real-world abductive NLI benchmark) of the four confusion probes used in this paper as perturbation-based interventions and detailed on the next page. Prompt-based interventions are shown at the top, and choice-based interventions are shown at the bottom. In aNLI, the prompt in an instance comprises two observations, and the candidate choices are the two hypotheses, of which exactly one is considered correct in the original unperturbed instance.}
\label{fig:confusion_probes}
\end{figure}

Before proceeding with working definitions of choice paralysis and prior bias, we introduce some basic formalism for placing the remainder of the paper in context. First, let us define an \emph{instance} $I=(p,A)$ as a pair composed of a \emph{prompt} $p$, and a set $A=\{a_1, a_2, ..., a_n\}$ of $n$ candidate \emph{choices}. We clarify the reasons for this terminology in \emph{Methods}, but intuitively, we use \emph{prompt} (rather than \emph{question}) because the input may not be a proper question\footnote{In general, commonsense benchmarks are NLU benchmarks, which \emph{may} involve QA, but do not have to. In some cases, the task is abductive reasoning, while in others, the task is NLI or even goal satisfaction, as in the case of the subsequently described Physical IQA benchmark.}. An example of such a case is provided in the center of Fig \ref{fig:confusion_probes}. 

Given an instance $I=(p,A)$, we assume that exactly one of the choices $\hat{a} \in A$ is \emph{correct}. Given a language representation model (such as one based on RoBERTa) $f$ that is designed to handle multi-choice NLI instances, we assume the output of $f$, given $I$, to be $(a', C)$, where $a' \in A$ is the model's predicted choice, and $C=\{c_1,c_2,..., c_n\}$ is a \emph{confidence set} that includes the model's confidence $c_i$ per candidate choice $a_i \in A$. We denote the variance of $C$ (calculated per instance) as $\sigma_C > 0$. We say that $I$ is \emph{perturbed} either if $p$ is changed in some manner (including being assigned the `empty string') or if $A$ is modified through addition, deletion, substitution or any other modification of candidate choices. Finally, if a perturbation applied on $I$ results in the perturbed instance $I_p$ not having any \emph{theoretically correct} choice in response to the prompt\footnote{We discuss four specific types of perturbations or `confusion probes' used in our experimental study in the next section, but for present purposes, we note that these definitions and formalism apply regardless of the form of the perturbation itself.}, but $\hat{a}$ is still a candidate choice, we refer to $\hat{a}$ as the \emph{pseudo-correct choice}. An obvious example of when this occurs is a perturbation that `deleted' the prompt by assigning it the empty string. Since there is no prompt, none of the candidate choices are theoretically correct or incorrect. Assuming that $A$ was not modified, the pseudo-correct choice would be $\hat{a}$.

With these basic preliminaries in place, we define the prior bias of a multiple-choice NLI model with respect to a perturbed instance below:

\textbf{Definition 1 (Prior Bias).} \emph{Given an original instance $I=(p,A)$ (with correct choice $\hat{a} \in A$), a perturbation of that instance $I_p$ that does not have any correct choice, and a multiple-choice NLI model $f$ that respectively outputs $(a', C)$ and $(a'_p, C_p)$ given $I$ and $I_p$, we define $\sigma_c'$ as the prior bias of $f$ with respect to $I_p$.}

Note that Definition 1 above only applies to perturbed instances where there is no correct choice (although there may potentially be a pseudo-correct choice $\hat{a} \in A$, depending on the specific type of perturbation applied). Definition 1 can also be generalized to quantify the prior bias of $f$ on an NLI benchmark (with respect to a specific perturbation) by aggregating $\sigma_c'$ across all perturbed instances in the benchmark. Only if the null hypothesis that $\sigma_c'=0$ cannot be rejected at a given level of confidence can we say with statistical certitude that $f$ does not have prior bias on that benchmark, with respect to the applied perturbation.

Next, using the same formalism, we can define choice paralysis in the context of multiple-choice NLI models:

\textbf{Definition 2 (Choice Paralysis).} \emph{Given an original instance $I=(p,A)$ (with correct choice $\hat{a} \in A$), a perturbation of that instance $I_p=(p_p,A_p)$ (with $|A_p| > |A|$, $\hat{a} \in A_p$, and $\hat{a}$ being the correct choice in response to prompt $p_p$), and a multiple-choice NLI model $f$, $f$ is said to have choice paralysis with respect to $I_p$ if the confidence of $f(I_p)$ in $\hat{a}$ is significantly lower than the confidence of $f(I)$ in $\hat{a}$. Denoting these two confidences respectively as $\hat{c}_p$ and $\hat{c}$, the magnitude of choice paralysis is given by $\hat{c}_p-\hat{c}$.}

Note that the direction of the subtraction matters i.e., $f$ can theoretically have negative choice paralysis whereby its confidence in the correct answer actually increases when a specific perturbation introduces a choice-set that is larger than the original choice-set. However, one important aspect that we note about Definition 2 is that $A_p$ does not have to be a super-set of $A$, although it is required to be larger, and at minimum, must contain the correct choice $\hat{a}$, similar to $A$. Furthermore, while there is no restriction on also perturbing the prompt $p$ (to a new prompt $p_p$), the perturbation must not be such that the theoretically correct answer changes. In practice, as we subsequently describe, our perturbation functions operate either at the level of the prompt, or choice-set, but not both. 


Finally, we note that, although both Definitions 1 and 2 impose some constraints on the types of perturbations that can be applied as interventions on the original instances in a benchmark, they can work with any perturbation functions that adhere to these constraints. For instance, as noted earlier, Definition 1 can be used to measure prior bias as long as there is no theoretically correct choice in the perturbed instance. While it may be possible to modify the definition to also measure prior bias if this were not the case, we leave such an expanded definition and its empirical validation to future research. Conversely, Definition 2 assumes that, even after the perturbation, the instance continues to contain a single theoretically correct choice in its (expanded) candidate choice-set $A$ in response to the prompt.

Since the definitions do not dictate specific perturbation functions, in order to conduct experiments, we need to devise one or more perturbation functions that enable us to quantify these phenomena for real-world multi-choice NLI benchmarks, and a sufficiently powerful language representation model that can handle not only the original benchmarks, but also their perturbed versions. Next, we describe the four perturbation functions used in this paper for studying these phenomena.

\subsection{Perturbation Methodology: Prompt-Based and Choice-Based Confusion Probes}

We designed a set of four perturbation functions, also called \emph{confusion probes}, that operate by systematically transforming multiple-choice NLI instances in four publicly available benchmarks, which have been widely used in the literature for assessing machine commonsense performance. These four benchmarks test the ability of a language representation model to select the best possible explanation for a given set of observations [aNLI \cite{anli, aNLI_link}], do grounded commonsense inference [HellaSwag \cite{hellaswag_link, hellaswag}], reason about both the prototypical use of objects and non-prototypical, but practically plausible, use of objects [PIQA \cite{piqa, piqa_link}], and answer social commonsense questions [SocialIQA \cite{socialiqa, siqa_link}].

Each of the four confusion probes intervenes either at the level of the prompt, or the candidate choices, but not both. As noted earlier, an instance comprises both a prompt and a candidate choice-set. The prompt may or may not be an actual `question', as understood grammatically. For example, as shown at the center of Fig \ref{fig:confusion_probes}, a single QA instance in the aNLI benchmark consists of two observations (the prompt) and two hypotheses (the choices or answers), of which one must be selected as being the best possible explanation for the given observations (abductive reasoning). In each of the four benchmarks, the structure of the instance is fixed, including the way in which the question is presented and the number of choices. The structure of instances in all four benchmarks, with an illustrative example each, is further detailed in \emph{Methods}.  

We designed two prompt-based interventions (\emph{No-Question} probe and \emph{Wrong-Question} probe), and two choice-based interventions (\emph{No-Right-Answer} probe and \emph{Choice Paralysis} probe), named intuitively and defined below. The \emph{No-Question} intervention has been used in many other NLP tasks. For example, \cite{kaushik2018much} used this probe to test the difficulty of reading comprehension benchmarks. The examples in these benchmarks are tuples consisting of question, passage, and answer. In their experiments, they analyzed the model's performance on various benchmarks when the test examples were provided with question-only or passage-only information (but not both). Similarly, in an NLI task, \cite{gururangan2018annotation, poliak2018hypothesis} re-evaluated high-performing models on hypothesis-only examples. In their experiments, models predicted the label (`entailment', `neutral' or `contradict') of a given hypothesis without seeing the premise. The \emph{No-Question} probe in our experiment is similar to the hypothesis-only model, which only provides models of multiple-choice choices without prompts. 

All our probes are visualized using an actual instance from a benchmark in Fig \ref{fig:confusion_probes}. Implementation details on how these benchmarks were manipulated to achieve these interventions are provided in \emph{Methods}, and can be used for replication. 

\begin{enumerate}
    \item {\bf No-Question probe:} This intervention probes the model by removing the prompt altogether and only retaining the candidate choice-set\footnote{In a slight abuse of terminology, and for reasons of maintaining compatibility with the QA literature where the use of the terms `prompt' and `instance' are non-standard, we use the (more specific) terms `question' and `answer' in the names of the probes, although in the main text we continue to use the (broader, but more accurate) terminology introduced in the early part of the paper.}.  
    \item {\bf Wrong-Question probe:} Similar to the No-Question probe, this probe retains the original candidate choice-set for an instance, but `swaps' the original prompt in an instance with a prompt from another instance (in the same benchmark) that is not relevant to any of the answers.
    \item {\bf No-Right-Answer probe:} This probe retains the prompt and all \emph{incorrect} choices in an instance, but replaces the correct choice with a choice from another instance's choice-set. The model is thus presented with a prompt but no correct choice in response to the prompt (among the presented choices). Note that, in the Wrong-Question probe, the candidate choices are completely unrelated to the corresponding prompt whereas in the No-Right-Answer probe, the non-substituted choices are often found to be semantically related to the prompt\footnote{For example, in Fig \ref{fig:confusion_probes}, the original prompt contains two Kat-related observations. Following the Wrong-Question intervention, the instance is given two Tom-related observations, which are completely unrelated to the two original Kat-related hypotheses. In contrast, following the No-Right-Answer intervention, the observations remain the same, but the original (correct) Kat-related hypothesis is substituted by a Jim-related hypothesis. Semantically speaking, the substituted hypothesis is less related to the observations than the non-substituted (but still incorrect) one, which mentions Kat.}. Theoretically, since all the choices are incorrect (in either probe), an ideal model would not exhibit higher prior bias for this probe, compared to the Wrong-Question probe, by getting confused by the misleading choice; however, we experimentally test whether this is indeed the case for the multiple-choice NLI system.
    \item {\bf Choice Paralysis probe:} Different from the three probes above, this probe is designed to test choice paralysis, as defined earlier. This probe `expands' the set of choices significantly using a systematic methodology that requires a parameter $n$ specifying the total number of choices to present to the model. $n$ must be higher than the number of choices originally presented in an instance per prompt. We rely on one of two sampling mechanisms to decide how the `incorrect' choices (the choices other than the correct choice) are selected from the other instances in the benchmark. The first of these is simple random sampling, while the second is `heuristic' sampling. In each of these, the first common step is to eliminate the original incorrect choices, followed by sampling the $n-1$ incorrect choices from the set of \emph{all correct choices} from the \emph{other} instances. When using simple random sampling, we sample the $n-1$ choices randomly, and with uniform probability, from that set. When using heuristic sampling, we select the $n-1$ choices that have greatest cosine similarity to the prompt in the RoBERTa sentence-embedding space. The latter, while not explicitly designed to be adversarial, is expected to lead to more confusion for the model as it selects answer choices that are somewhat related to the prompt in the embedding space, but that are still incorrect.     
\end{enumerate}


We note that a unique aspect of all of our probes (with the exception of choice paralysis) is that the performance, as measured using \emph{pseudo-accuracy} (defined simply as the fraction of test-set instances where the model picked the pseudo-correct choice), should ideally \emph{decline} post-intervention. This ideal performance (which is the same as random performance) occurs if the model has no prior bias. This makes the first three confusion probes different from other such experiments evaluating language models. For example, we would ideally want a model to pick among choices randomly when it is not presented with a prompt (\emph{No-Question} probe). However, our probe and methodology is able quantify the magnitude (and statistical significance) of this bias.
Finally, for all probes and benchmarks, we conduct, and interpret the results of, each experiment not only by comparing the model's post-intervention pseudo-accuracy performance to its original performance (on the unperturbed instances), but also by comparing its pseudo-accuracy to the \emph{expected} performance that would be achieved by a system without any prior bias. 

\section{Methods}
\subsection{Evaluation Datasets}
\begin{figure}[ht]
\centering
\includegraphics[width=10cm]{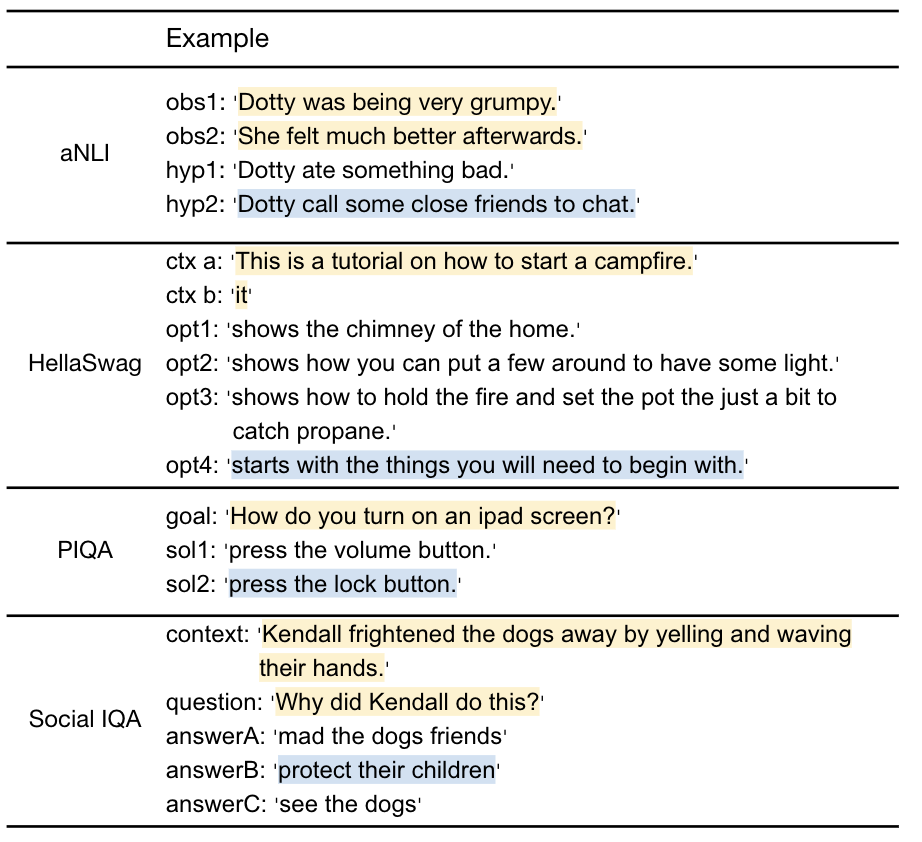}
\caption{Instances (prompt and candidate choice-set) from the four commonsense benchmark datasets used for the experimental study herein. The prompt is highlighted in yellow, and the correct choice is in blue.}
\label{fig:benchmarks}
\end{figure}

The four benchmarks used in the experimental study are described below, with references for further reading. We also provide a representative example in Fig \ref{fig:benchmarks}.  We emphasize that an instance is the combination of the prompt and the candidate choice-set, and only if the prompt is an actual question, should the instance technically be thought of as a QA instance. In the general case, each instance should be thought of broadly as testing natural language understanding. In the rest of the discussion, we continue using the proper terms `instance', `choice' and `prompt' (rather than the somewhat inaccurate terms `QA instance', `question' and `answer', respectively) to refer to these concepts.


\begin{enumerate}
    \item {\bf aNLI (Abductive Natural Language Inference):} Abductive Natural Language Inference (aNLI) \cite{anli, aNLI_link} is a commonsense benchmark dataset designed to test an AI system's capability to apply everyday abductive reasoning to deduce possible explanations for a given set of observations. Formulated as a binary-classification task, the goal is to pick the most plausible explanatory hypothesis given two observations (from narrative contexts).
  The combination of the two observations (provided as input in a given instance) is considered the prompt. The benchmark contains 169,654 instances in the training set and 1,532 instances in the development (\emph{dev.}) set, which is used as `test set' in experiments. Compared to human performance of 0.93, the highest performance of a language model (at the time of writing) is 0.90, which is achieved by DeBERTa (Decoding-enhanced BERT with disentangled attention), produced by Microsoft Dynamics 365 AI.
    
    \item {\bf HellaSwag:} HellaSWAG \cite{hellaswag_link, hellaswag} is a dataset for studying grounded commonsense inference. It consists of 49,947 multiple-choice instances about `grounded situations' (with 39,905 instances in the training set and 10,042 instances in \emph{dev.} set). Each prompt comes from one of two domains--Activity Net or wikiHow--with four candidate choices about what might happen next in the scene. The correct choice is the (real) sentence for the next event; the three incorrect choices are adversarially generated and human-verified, ensuring a non-trivial probability of `fooling' machines but not (most) humans. Each HellaSwag instance provides two \emph{contexts} as the prompt. UNICORN \footnote{\url{https://leaderboard.allenai.org/anli/submission/bsd2tf9bvhc9b55n46eg}}, a model based on T5, achieves the current highest performance (0.94) of models on this benchmark, which approaches human performance (0.96).
    
    \item {\bf PIQA:} Physical Interaction QA (PIQA) \cite{piqa, piqa_link} is a novel commonsense QA benchmark for na{\"i}ve physics reasoning, primarily concerned with testing machines on how humans interact with everyday objects in common situations. It tests, for example, what actions a physical object `affords' (e.g., it is possible to use a cup as a doorstop), and also what physical interactions a group of objects afford (e.g., it is possible to place an computer on top of a table, but not the other way around). The dataset requires reasoning about both the prototypical use of objects (e.g., glasses are used for drinking) but also non-prototypical (but practically plausible) uses of objects. There are 16,113 instances in PIQA's training set and 1,838 instances in its \emph{dev.} set. The goal in every PIQA instance is the prompt in our experiments. Compared to human accuracy (0.95), the machine's best performance is 0.90, which is also achieved by UNICORN.
    \item {\bf Social IQA:} Social Interaction QA \cite{socialiqa, siqa_link} is a QA benchmark for testing social common sense. In contrast with prior benchmarks focusing primarily on physical or taxonomic knowledge, Social IQA is mainly concerned with testing a machine's reasoning capabilities about people's actions and their social implications. Actions in Social IQA span many social situations, and candidate choices comprise of both human-curated answers and `adversarially-filtered' machine-generated choices. Social IQA contains 33,410 instances in its training set, and 1,954 instances in its \emph{dev.} set. While Social IQA separates the context from the question, the two collectively constitute the prompt, in keeping with the terminology mentioned earlier. We note that both human- and machine-performance on Social IQA are slightly lower than other benchmarks. Specifically, human accuracy on Social IQA is 0.88, with UNICORN achieving an accuracy of 0.83.
\end{enumerate}

\subsection{RoBERTa-based Model}

Transformer-based models have rapidly emerged as state-of-the-art in the natural language processing community, both for specific tasks like question answering, but also for deriving `contextual embeddings'. BERT is a bi-directional transformer that can be pre-trained over a lot of unlabeled textual data to learn a language representation model. This model can, in a second step that does not typically use as much data as the first step, be \emph{fine-tuned} for specific machine learning tasks.

RoBERTa is a more optimized re-training of BERT that removes the Next Sentence Prediction task from BERT's pre-training, while introducing dynamic masking so that the masked token changes during the training epochs. Larger batch-training sizes were also found to be more useful in the training procedure. Unlike a more recent model like GPT-3, a pre-trained version of RoBERTa is fully available for researchers to use and can be fine-tuned for specific tasks \cite{liu2019roberta}.  

Unsurprisingly, many of the systems occupying a significant fraction of top leaderboard\footnote{\url{https://leaderboard.allenai.org/}} positions for the commonsense reasoning benchmarks described earlier are based on RoBERTa (or some other optimized BERT-based model) in some significant manner. All experiments in this paper use a publicly available \emph{RoBERTa Ensemble} model \footnote{\url{https://github.com/isi-nlp/ai2/tree/base}} that was not developed by any the authors, either in principle or practice, and that can be downloaded and replicated very precisely. It is also worth noting that even recent models that have superseded RoBERTa (such as T5) on the benchmarks are based on transformers as well. In the \emph{Discussion} section, we showed that (at least one of) the newer models may also exhibit biases when subject to the full set of confusion probes. 

The RoBERTa Ensemble model is fine-tuned on each benchmark's respective training set and evaluated on their \emph{dev.} set to test the model's performance. Each such trained model was verified to achieve over 80\% performance (on average) over the four benchmarks, and performance is not substantively different from the state-of-the-art performance recorded on the current leaderboards. This fine-tuned model will be re-used in the experiments below to evaluate post-intervention instances after applying the different (previously described) probes. We do not re-fine-tune the model post-intervention.We also note that the model continues to work (i.e., successfully makes a prediction) even for the \emph{Choice Paralysis} confusion probe, where there is a mismatch between the training and evaluation set formats, due to instances in the evaluation set having more candidate choices compared to the training set. The only syntactic modification that was needed was to input to the model the number of choices in the candidate choice-set (presented to it per instance in the test phase), from which it needed to predict the correct choice. The model itself did not have to be re-fine-tuned. 

\subsection{Instance Perturbations Using Confusion Probes}
Earlier, we described the four confusion probes (\emph{No-Question}, \emph{Wrong-Question},\emph{No-Right-Answer}, and \emph{Choice Paralysis}) that we rely on for our experimental study. Below, we provide additional details relevant to their setup and implementation for the RoBERTa Ensemble model: 
\begin{enumerate}
\item {\bf No-Question}: For each instance, we remove the prompt and make the RoBERTa-Ensemble model select a choice from its original choice-set without any contexts or prompts. Note that the RoBERTa-Ensemble model is capable of of accepting an empty string as prompt. 
\item {\bf Wrong-Question}: For each instance, we replace the original prompt with a \emph{mismatched} prompt (called the `pseudo-prompt'). The pseudo-prompt for a given instance is an actual prompt from another randomly selected instance in that benchmark. No change is made to the choice-set. While there is a very small probability that the pseudo-prompt may be correctly answered by a choice from the unmodified choice-set, in practice, we could find no such cases when we randomly sampled and manually checked (post-intervention) 25 instances from each benchmark. As a further robustness check, we conducted another experiment where, instead of randomly sampling an instance from which the pseudo-prompt was selected, we only considered sampling from instances where the prompt did not share any words with the original prompt. The experimental results were not found to change appreciably compared to the simpler replacement protocol described above. Hence, we only report those results for that protocol. Furthermore, to account for randomness, we repeat the experiment five times (per benchmark), and report average performance with standard errors. In the ideal situation (where the model would not get confused and exhibit prior bias), the model should still randomly choose an option from the choice-set, similar to its ideal response on the \emph{No-Question} probe.

\item {\bf No-Right-Answer}: For this probe, the choice-set is modified rather than the prompt. Namely, each instance's correct choice is substituted with a mismatched choice (called the `pseudo-choice'). Similar to the \emph{Wrong-Question} replacement protocol, we randomly sample an instance and substitute the `correct' choice of the given instance with the correct choice of the randomly sampled instance. We conduct similar robustness checks (and also manual checks) as with the \emph{Wrong-Question} protocol, but again found the results to be similar to that of the described protocol. Hence, only those results are reported. We also account for randomness through five experimental trials. Without a correct answer for the prompt, the ideal model would be expected to randomly pick a choice from all the presented (and all incorrect) choices.

\item {\bf Choice Paralysis}: The prompts of instances are kept unchanged, but the choices are expanded significantly. There are two ways to extend the choice-set. Assuming that the number of options for each instance is a parameter $n$ (with $n=5, 10, 15$ in the experiments reported herein), for each target instance, we randomly pick $n-1$ other instances at first. The correct answers of these $n-1$ instances (one per instance), together with the correct answer of the target instance, constitute the $n$ choices. Instead of randomly picking the $n-1$ instances, we also experimented with a heuristic approach that picks out $n-1$ instances that have the prompts most similar to the prompt of the target instance. The similarity between two prompts is computed by calculating the cosine similarity between the sentence embeddings of the prompts. We randomly pick out 50 modified instances from the original \emph{dev.} set as a new \emph{dev.} set for evaluation and repeat this evaluation five times to account for randomness. A smaller \emph{dev.} set is considered for this experiment due to the significant increase in computational time incurred per prompt (for the language representation model) when the choice-set is expanded. We report both the average and standard errors. Although the choices have been extended, the original prompt and the correct answer (for the original prompt) are retained; hence, an ideal model that does not suffer from choice paralysis is expected to choose the correct answer. 
\end{enumerate}

\section{Results}
\subsection{Question-Based Interventions}
\subsubsection{No-Question Probe}

\begin{figure}[ht]
\centering
\includegraphics[width=10cm]{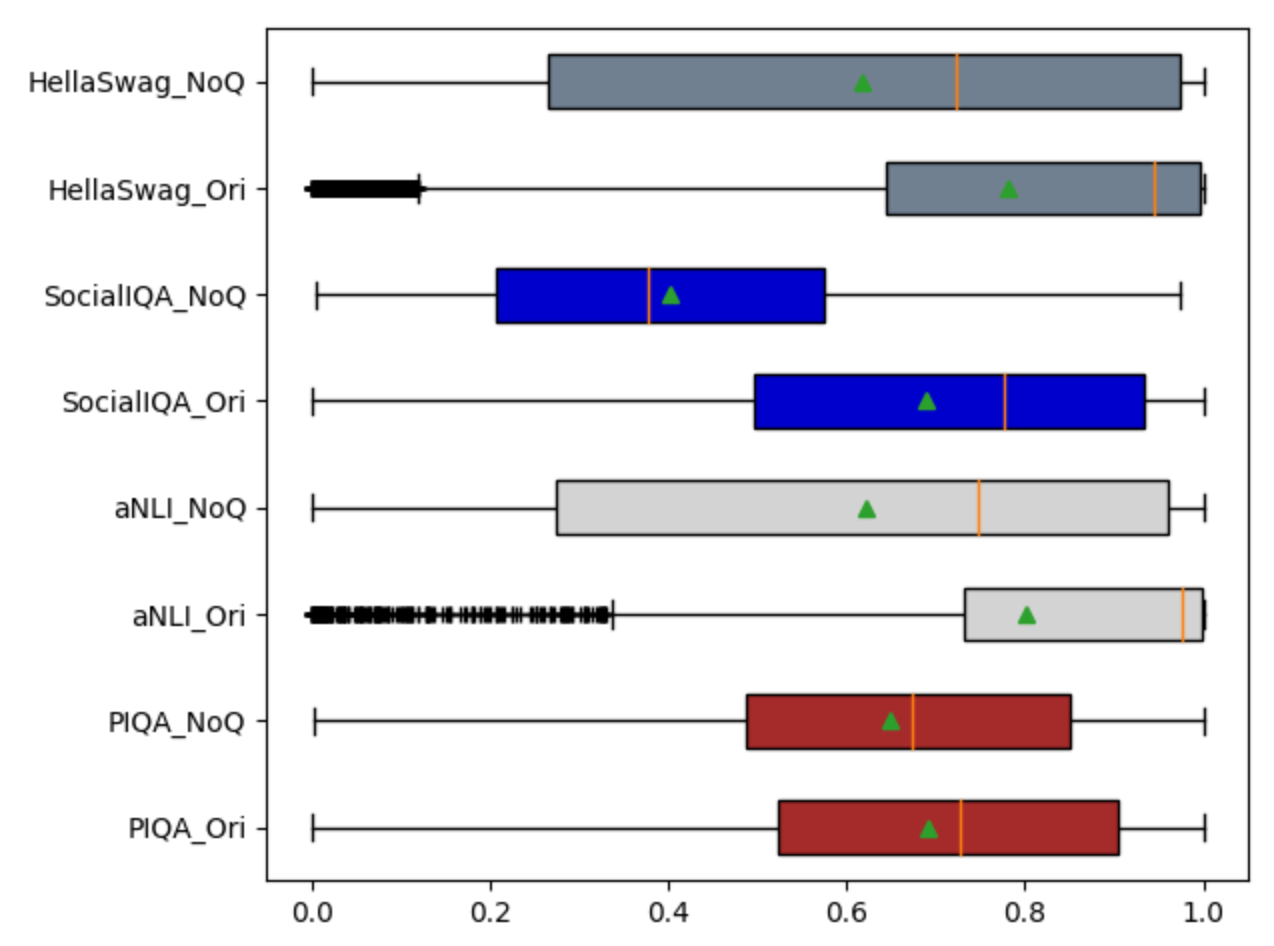}
\caption{A boxplot summarizing the confidences of correct (and `pseudo-correct') options before (denoted using \emph{\_Ori}) and after the \emph{No-Question} intervention. The orange line and green triangle respectively represent the median and mean. We use '+' markers in the plot to indicate outliers.}
\label{fig:NoQ}
\end{figure}


We report the differences between confidence distributions of correct and `pseudo-correct' options before and after the intervention (namely, removing the prompt) respectively in Fig \ref{fig:NoQ}. Along with the actual distribution, we also report the median and mean of the confidences of the correct and  `pseudo-correct' options for all four benchmarks. Ideally, we expect the post-intervention confidence of the pseudo-correct option to equal the reciprocal of the (benchmark-specific) number of choices per instance. We refer to this as the \emph{bias-free} confidence. In other words, a system that behaves ideally would sample a choice from the uniform probability distribution over all the available choices when the prompt is not available. The reason is that, after removing the prompt, there is no correct answer anymore. While some prior bias is expected, as discussed briefly in \emph{Background}, the extent of the bias is an open question that can only be observed empirically.

The figure shows that, while both the mean and median confidence of the pseudo-correct option decreases in each of the four benchmarks compared to the pre-intervention confidence of that (then correct) option, the extent of the decline depends on the benchmark and never equals the bias-free confidence, which equals 0.5 for PIQA and aNLI, 0.25 for HellaSwag and 0.33 for SocialIQA (reported also in Table \ref{tab1} in the next experiment). We find that the pre-intervention confidence distribution obtained by RoBERTa on PIQA in particular is very similar to the post-intervention distribution. On SocialIQA, in contrast, the difference is more marked, and the post-intervention mean confidence almost achieves the bias-free value of 0.33. Therefore, RoBERTa is more likely to randomly choose an option for the SocialIQA benchmark after the prompt is removed, while there is significant prior bias in PIQA.

The \emph{dispersion} of post-intervention confidence also differs from expectations that would hold in a bias-free setting. On SocialIQA and PIQA, RoBERTa obtains a slightly smaller inter-quartile range of post-intervention confidences; however, considerably larger inter-quartile ranges are observed on HellaSwag and aNLI, compared to the pre-intervention dispersion. The confidences of `pseudo-correct' options in HellaSwag and aNLI are therefore more variable than in the other benchmarks. Thus, for some instances, RoBERTa needs the prompt to guide it toward the correct answers, but for others, it can directly make decisions by only looking at the options and without requiring any prompt. Put together, the results suggest fairly strong prevalence of prior bias, with SocialIQA serving as the lone exception (and with considerable dispersion of its own). 



\subsubsection{Wrong-Question Probe}


Key findings for the \emph{Wrong-Question} probe can be found in Table \ref{tab1}. For reference, we show in Column 2 the `bias-free' pseudo-accuracy (i.e. without any prior bias, as discussed toward the end of \emph{Background}), again, is equal to the reciprocal of the (benchmark-specific) number of choices per instance. This concept is defined similarly as in the previous experiment. Note that this is technically also equal to the bias-free confidence in the pseudo-correct choice. In contrast, in Column 4, we report the actual (average) pseudo-accuracy achieved on each of the four benchmarks using the RoBERTa model, following the \emph{Wrong-Question} intervention. 
\begin{table}[]
\centering
\begin{tabular}{|l|p{1.4in}|p{1.0in}|p{1.4in}|}
\hline
{\bf Dataset}   & {\bf Bias-free pseudo-accuracy / confidence} & {\bf Average confidence difference} &  {\bf Actual pseudo-accuracy (+/- std. err.)} \\ \hline
PIQA      & 0.5                       & 0.015            & 0.72 (+/- 0.0014)           \\ \hline
aNLI      & 0.5                       & 0.044            & 0.59 (+/- 0.0024)            \\ \hline
SocialIQA & 0.33                      & 0.022            & 0.40 (+/- 0.0009)             \\ \hline
HellaSwag & 0.25                      & 0.060            & 0.61 (+/- 0.0081)            \\ \hline
\end{tabular}
\caption{A summary of results following the \emph{Wrong-Question} intervention.}\label{tab1}
\end{table}

Consistent with what we found in Fig \ref{fig:NoQ}, RoBERTa achieves a high pseudo-accuracy post-intervention on PIQA, and closer-to-ideal pseudo-accuracy on SocialIQA. Therefore, on PIQA the model is clearly susceptible to this kind of confusion, possibly because of a strong prior bias. While not as extreme as SocialIQA, the model does show a marked decrease in pseudo-accuracy on the HellaSwag and aNLI benchmarks, compared to the original pre-intervention accuracy. Standard errors on HellaSwag and aNLI are also higher than on PIQA and SocialIQA, which is a direct consequence of the confidences of pseudo-correct options on these two benchmarks being more variable (than PIQA and SocialIQA). In all cases, results are significantly different, at the 99 percent confidence level or higher, compared to both the ideal pseudo-accuracy, as well as the pre-intervention accuracy.  The latter is encouraging, but expected, and supports the intuition that the prompt \emph{matters} for the model but the former suggests that it matters less than it should.  

In addition, the average confidence difference between the \emph{No-Question} probe and the \emph{Wrong-Question} probe in each benchmark is not only positive (Column 3 in Table \ref{tab1}), but we have verified all of them to be significantly greater than 0 with at least 99 percent confidence. The result implies that the appearance of the wrong prompt makes the model start to `doubt' the pseudo-correct option compared to the \emph{No-Question} probe, and to reconsider which option is more correct for the current mismatched prompt (following the \emph{Wrong-Question} intervention). On average, we have also verified the confidence of the model in the pseudo-correct option to be significantly different from the pre-intervention confidence in that (then correct) option. We also confirmed that this average pseudo-correct confidence is significantly different from the  ideal confidence noted in Column 2. 

It bears noting that even in the best case (SocialIQA), although RoBERTa's pseudo-accuracy is significantly closer to the ideal compared to the other benchmarks, it is still significantly different. Therefore, the assumption that the model needs to be presented with an actually correct option is a powerful one that should not be underestimated in practice. When the options are all incorrect, the model does \emph{not} just randomly or uniformly pick one out as its choice. There is clear evidence of a prior bias, a finding also supported earlier by the \emph{No-Question} probe. 




\subsection{Choice-Based Interventions}
\subsubsection{No-Right-Answer Probe}

\begin{figure}[ht]
\centering
\includegraphics[width=10cm]{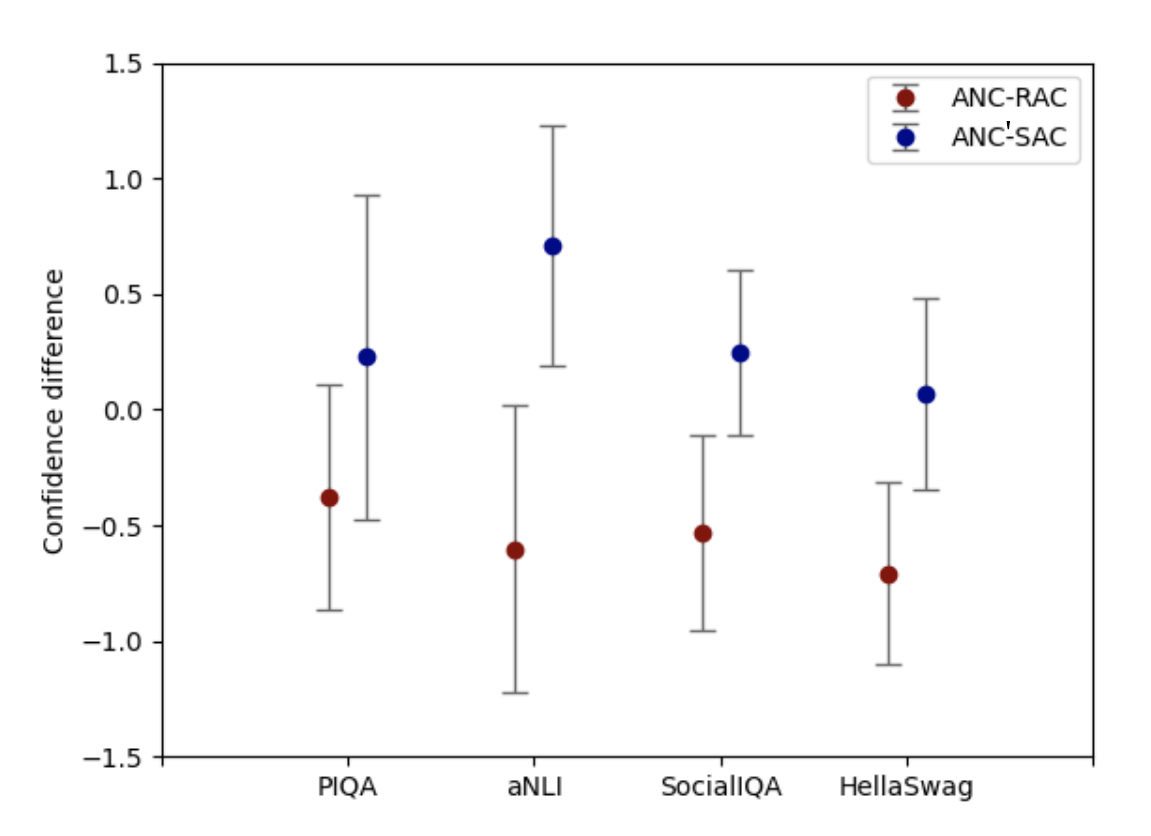}
\caption{Mean of confidence differences (with error bars) for each benchmark before and after the \emph{No-Right-Answer} intervention. ANC, ANC', RAC and SAC are defined in the text. The error bars represent standard errors of confidence differences, ranging from 0.37 to 0.75.}
\label{fig:NoA}
\end{figure}

Fig \ref{fig:NoA} summarizes results on the \emph{No-Right-Answer} probe using two key statistics. Before describing these statistics, we define some supporting measures. Specifically, let ANC and ANC' respectively represent the average confidence of the model in the \emph{non-substituted} options pre- and post-intervention. For benchmarks such as PIQA and aNLI, this is the confidence of the lone non-substituted option, since they only offer two choices per instance. For the other two benchmarks, where there are more than two choices, we average the confidences of all incorrect non-substituted choices. In contrast, RAC represents the confidence of the real correct option \emph{before} the intervention, while SAC represents the confidence of the substituted option after swapping out the correct option (following the intervention). 

Using the measures above, we can now define the two key statistics $ANC'-SAC$ and $ANC-RAC$. The former is a measure of how biased the model is toward non-substituted incorrect options compared to the substituted incorrect option. Note that the non-substituted options tend to have more \emph{surface similarity} to the prompt (because of the way the benchmarks were conceived and designed, in order to be reasonably challenging for QA models), compared to the substituted option, even though they are all incorrect. The $ANC-RAC$ measure is instead a pre-intervention measure, and measures the confidence gap of the model between the non-substituted choices and the original correct choice. Ideally, $ANC-RAC$ should be -1.0, and we do indeed find that the model tends toward this number for almost all four benchmarks. 

In contrast, for the $ANC'-SAC$ measure, the model should tend toward 0 when only the correctness of an option, rather than surface similarity to the prompt, matters. In fact, given the positive bias of $ANC'-SAC$ in Fig \ref{fig:NoA}, it is clear that RoBERTa tends to choose one of the non-substituted options. Among  the four benchmarks, RoBERTa obtained the highest $ANC'-SAC$, as well as relatively low $ANC-RAC$ (and, consequently, the greatest \emph{difference} between the $ANC'-SAC$ and $ANC-RAC$ measures) on the aNLI benchmark. Interestingly, we observe higher pre-intervention standard errors for the $ANC-RAC$ measure for aNLI, in contrast with that of PIQA, where the post-intervention spread is higher. Furthermore, for PIQA, the post-intervention and pre-intervention confidence spreads overlap much more than for other benchmarks.

Perhaps the most interesting qualitative takeaway from the results is that surface similarity between prompt and answer can clearly matter a lot, even when the answer is wrong. For difficult questions or particularly creative (and correct) answers to such questions, this bias may prove to be problematic. As we note in \emph{Discussion}, this problem also exists in the newer QA models that rely on billions of parameters to achieve better performance.


\subsubsection{Choice Paralysis Probe}

\begin{figure}[ht]
\centering
\includegraphics[width=12cm]{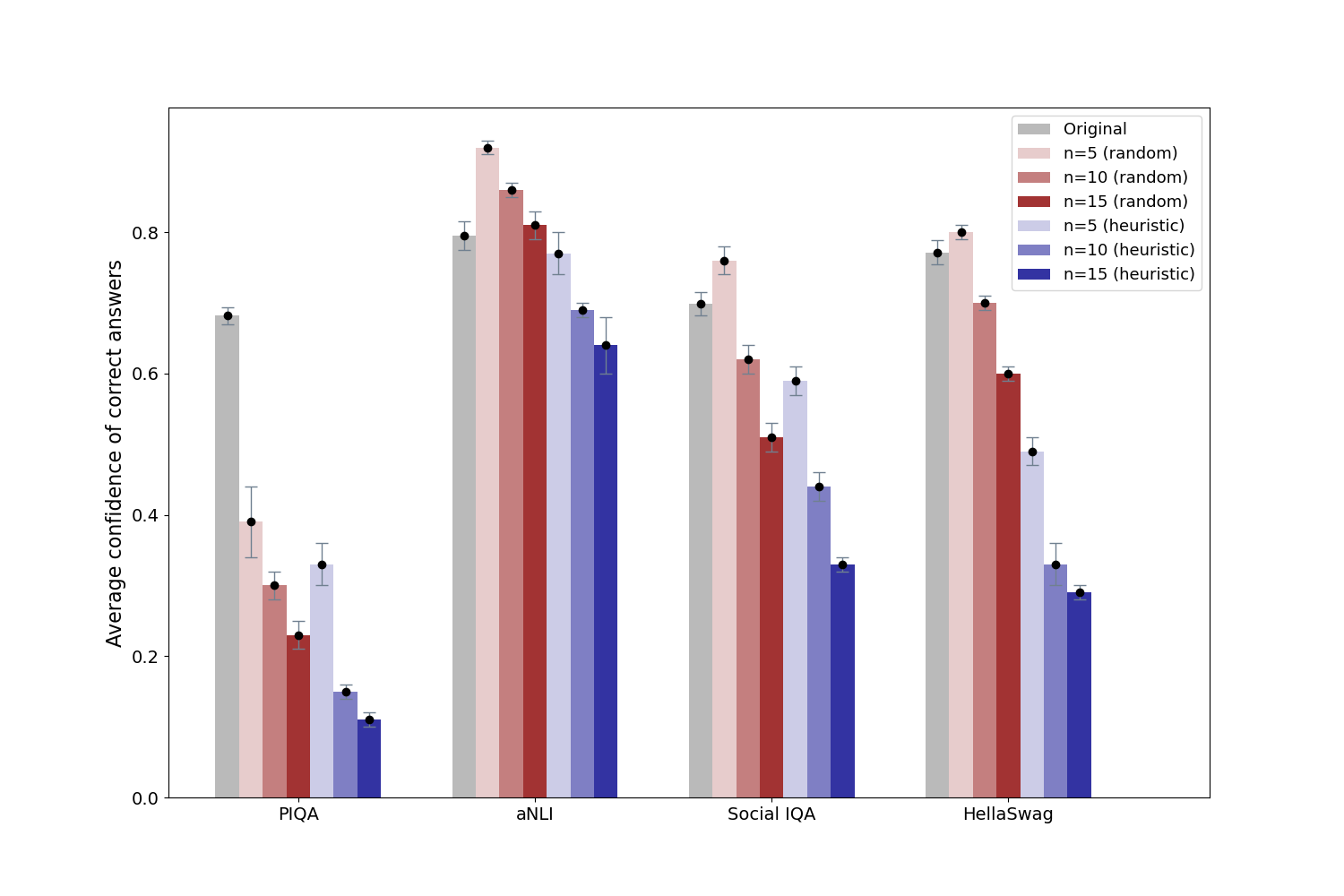}
\caption{The average confidence of RoBERTa in the correct option, with standard errors, following the \emph{Choice Paralysis} intervention. Results are displayed for different values of $n$ and choice-sampling methodologies (\emph{random} and \emph{heuristic}). The grey bar is the average confidence of the correct answer in the original \emph{dev.} set before the intervention). The error bars range from 0.01 to 0.07.}
\label{fig:choicePara_conf}
\end{figure}

\begin{figure}[ht]
\centering
\includegraphics[width=12cm]{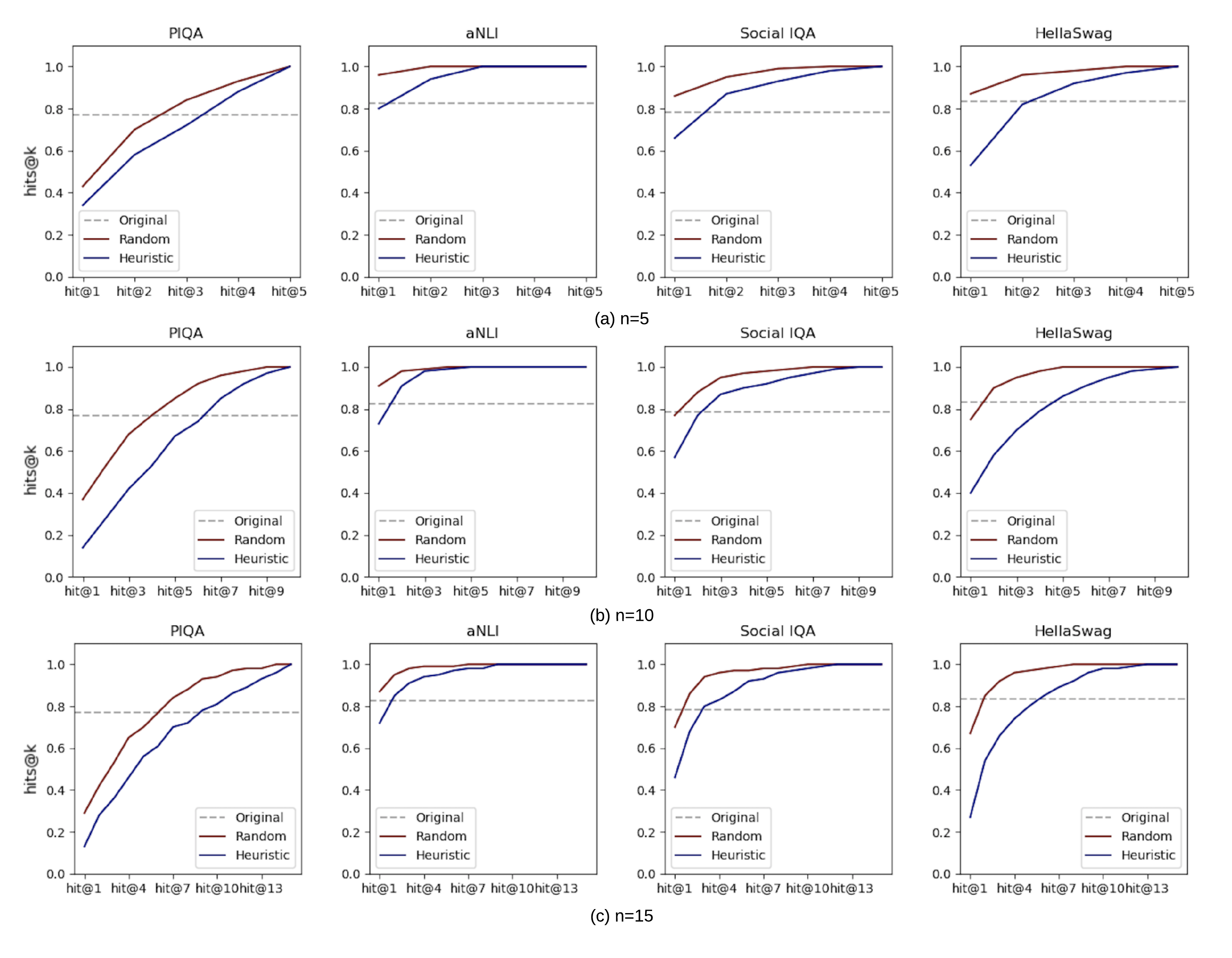}
\caption{The hits@k performance of the model following the \emph{Choice Paralysis} intervention. The dashed grey line in each sub-plot shows the original performance (accuracy) on each benchmark. As in Fig \ref{fig:choicePara_conf}, $n$ represents the total number of choices following the intervention, results are similarly illustrated for both choice-sampling methodologies.}
\label{fig:hit@k}
\end{figure}


Fig \ref{fig:choicePara_conf} illustrates our key findings when intervening using the \emph{Choice Paralysis} probe. In principle, a language model like RoBERTa should be able to choose the correct answer no matter how many choices are provided per instance, and the confidences of correct answers should be close to their pre-intervention confidence. In practice, we expect some loss in both accuracy and confidence, but similar to the earlier probes, the extent of this loss can only be determined empirically.  In Fig \ref{fig:choicePara_conf}, however, we find, interestingly enough, that when $n=5$ and the sampling methodology is random, RoBERTa is able to achieve \emph{better} performance on three of the four benchmarks (PIQA is the lone exception). This suggests that, when a random wrong answer is inserted as a choice, the model is better (with higher confidence) able to pick out the right answer on most of the benchmarks. 

When the sampling is heuristic, which is more adversarial since we are deliberately trying to confuse the model with an option that has greater chance of being more related to the prompt and to the other answers, $n=5$ leads to significant decline in performance for all benchmarks except aNLI (where there is a decline, but is not significant). In fact, the results suggest that aNLI is the `easiest' benchmark for RoBERTa when applying this specific probe, since it is the only benchmark where it is able to stay within a 20\% margin of its original performance even in the most aggressive setting ($n=15$, with heuristic sampling). Even so, the results clearly illustrate that even on this benchmark, RoBERTa is not immune from the choice paralysis problem indefinitely. 

 In general, except for $n=5$ with random sampling, the result is an expected one across all four benchmarks: average  post-intervention confidence on each of four benchmarks shows a decreasing linear-like dependence on the number of choices per instance, when random sampling is used. When $n$ goes from 5 to 10, and then 10 to 15, the model shows similar decrease, on average, in the confidence of the correct option. Results for the heuristic sampling methodology are much more straightforward, with RoBERTa exhibiting confusion consistently and to a more extreme extent than with the random sampling (keeping $n$ fixed).


One aspect of the results in Fig \ref{fig:choicePara_conf} is that, as the number of options expand, only taking the confidence of the correct option into account (or using the accuracy metric) to understand the confusion properties of the language model, may be too harsh. An alternate way to understand choice paralysis is to use the \emph{hits@k} metric. This metric does not use the confidence directly, but instead ranks the options in decreasing order of confidence. With this ranking in place, and given a value for $k$, the \emph{hits@k} for an instance is 1 if the correct answer falls within the top $k$ items on the ranked list.   

Since this metric clearly depends on $k$, we vary $k$ for each benchmark and experimental setting (using the different values of $n$ and the two sampling methodologies) and plot the results in Fig \ref{fig:hit@k}.  Consistent with earlier findings in Fig \ref{fig:choicePara_conf}, we find in Fig \ref{fig:hit@k} that RoBERTa continues to perform reasonably well relative to its original performance (the dashed horizontal line in each sub-plot), after \emph{random} sampling intervention, and its hits@2 accuracy is higher than the original accuracy on most benchmarks except when $n<=15$. PIQA is again an exception, just as in Fig \ref{fig:choicePara_conf}. For PIQA, the model needs to output 3 predictions (after a 5-choice intervention) before the correct option is covered at the same rate as its original accuracy. 

Once again, we find that the heuristic sampling methodology causes much more severe confusion for the model. RoBERTa starts to need more top ranked predictions, especially on PIQA and HellaSwag, to reach the original accuracy. On PIQA, the model may need to see more than half the options ($k=n/2$) before its hits@k accuracy equals the original accuracy. While not as extreme as PIQA, on HellaSwag, RoBERTa typically needs to make six predictions to maintain its original performance in the $n=15$ experimental setting.    

An additional important point that bears noting is that, as we expand the number of options $n$, the time-cost for answering a prompt exhibits an increasing near-linear relationship in $n$. For example, on PIQA, the average time that RoBERTa took to answer 50 instances went from 30 minutes to 98 minutes when presented with 15 options, compared to when it was presented with 5 options. As we briefly note in \emph{Discussion}, the time complexity can be more extreme with newer transformer-based models.  

\subsection{Confidence Calibration}
Recent studies have suggested that the BERT-based models can be overly confident in their predictions \cite{wallace2019a, ribeiro2018}. We use MaxProb as a calibration technique to distinguish pre- and post-intervention instances \cite{hendrycks2016baseline, kamath2020selective, zhang2021knowing}. As described below, MaxProb makes this distinction by using the probability assigned by the underlying multiple-choice NLI system to the most likely (i.e., highest-probability) prediction among the candidate choices. If it decides that an instance is post-intervention, it recommends that the NLI system abstain from answering. Previously, MaxProb was found to give good confidence estimates on multiple-choice benchmarks.

To train MaxProb, we randomly selected 50\% instances from a benchmark's dev. set, perturbed these instances using a probe\footnote{Note that we only test No-Question, Wrong-Question, and No-Right-Answer probes here.}, and used the mixture of original and perturbed instances as a training set. The average MaxProb over this set is treated as a threshold, with the other 50\% instances composed of an evaluation set. In the evaluation set, the instances are perturbed using the same probe. Only if the MaxProb of an instance is higher than the learned threshold, the instance will be predicted as an original (i.e., non-perturbed) instance. The accuracy of MaxProb is defined as the proportion of instances correctly predicted as pre- or post-intervention instances. 

Table \ref{table: maxprob} shows the per-benchmark learned thresholds using different confusion probes. Social IQA had the lowest thresholds (among all benchmarks) on all three confusion probes; however, its average learned MaxProb threshold\footnote{As mentioned before, RoBERTa should be equally confident about each of its candidate choices on a post-intervention instance if there is no prior bias. Hence, ideally, the MaxProb on a given benchmark should just be the reciprocal of the number of candidate choices (per instance, in that benchmark).} is still 0.7 or higher. Because of the high threshold, MaxProb should help RoBERTa more easily abstain from answering post-intervention instances. However, when we used MaxProb to distinguish perturbed instances in the evaluation set, we found its accuracy to only be about 0.64 or below on most benchmarks (except on the SocialIQA No-Question probe). The evidence also suggests that the No-Right-Answer instances are the hardest cases for MaxProb, with the Wrong-Question instances being the easiest. Although we focused on MaxProb in these additional experiments,  future work could substitute MaxProb for other, more recently proposed selective prediction methods\cite{kamath2020selective, zhang2021knowing}.

\begin{table}[]
\centering
\begin{tabular}{|r|r|r|r|}
\hline
          & No-Question & Wrong-Question & No-Right-Answer \\ \hline
aNLI      & 0.88 (0.59)  & 0.86 (0.64)  & 0.92 (0.46)   \\ \hline
HellaSwag & 0.82 (0.58)   & 0.8 (0.61)   & 0.8 (0.62) \\ \hline
PIQA      & 0.76 (0.53)    & 0.76 (0.55)  & 0.81 (0.41)  \\ \hline
SocialIQA & 0.70 (0.70)    & 0.74  (0.60)   & 0.76 (0.55)\\ 
\end{tabular}
\caption{The learned MaxProb thresholds of different confusion probes on four benchmarks. The accuracy of MaxProb to distinguish pre- and post-intervention instances in each evaluation set is shown in brackets.}\label{table: maxprob}
\end{table}

\section{Discussion}

While the results described in the previous section clearly indicate non-trivial, and often significant, evidence of both prior bias and choice paralysis, they do not shed much  light on the potential \emph{causes} of these issues. For example, is fine-tuning a major driver of prior bias, and are larger models that were released subsequent to BERT and RoBERTa equally prone to prior bias and choice paralysis? While causal questions are fundamentally difficult to address using only a limited set of benchmarks, models and interventions, we provide a discussion in this section using auxiliary analyses. Specifically, we aim to illustrate the potential impact of fine-tuning by evaluating other models, including a non-fine-tuned version of RoBERTa, and we also evaluate non-BERT-based models to understand if the problem is specific to BERT's architecture. We also study if syntactic properties or `irregularities' in the benchmarks themselves may have contributed to the issues.

\subsection{Irregularities Analysis}
The evidence from the experimental study suggests that specific benchmarks used to fine-tune the model have some effect on the bias; hence, we begin by conducting an `irregularities analysis' i.e., by testing effects of label imbalance, the distribution of prompt-lengths, and the words-overlap between the prompt and the candidate answers, on our benchmarks, to confirm or exclude some causes for the bias.

Specifically, the label imbalance analysis aims at testing whether the first listed choice is labeled in a benchmark's dev. set as correct with a higher-than-random probability. Statistically, we found that labels tend to be evenly distributed in all four benchmarks' dev. sets, and hence, there is no label imbalance or `choice ordering' bias of any kind. For example, in the 2-option PIQA benchmark, the first option is the correct answer in 49.5\% (910 out of 1838) of the instances. In aNLI, the first option is the correct answer in 51\% (781 out of 1532) of the instances. In the 3-option Social IQA benchmark, 33.4\% and 33.6\% of the instances are labeled with  `answerA' and `answerB' as the correct answer, respectively, which again suggests near-random label distribution. Similarly, in the 4-option HellaSwag, the frequencies of the four options labeled as the correct answers are 25\% (each).

Similarly, we also calculated the average \emph{prompt lengths} (in terms of the number of words) in the two sets comprising the selected (by the multiple-choice NLI system)  and non-selected candidate choices to determine if there is some kind of a length bias. While we did find that the selected choices tended to be longer than the non-selected choices over all benchmarks, the difference was slight and not statistically significant. For example, in PIQA, the average length of selected candidate answers is 19.24 words (with 95\% confidence interval of [18.38, 20.11]), while the average length of non-selected candidate answers is 18.43 words. In Social IQA, the average length of selected candidate answer is 3.72 words, (with 95\% confidence interval of [3.62, 3.82]), while the average length of non-selected candidate answers is 3.69 words. The difference was not found to be significant at the 95\% confidence level.

Finally, we analyzed the words-overlap between the candidate choices and the prompt, by grouping all candidate choices into two sets (selected and non-selected), similar to the grouping employed in the analysis above. We found that most of the overlapping words between the prompts and choices (in either set) were \emph{stopwords}. Indeed, when we calculated the Pearson's correlation between the word frequency distributions over the selected and non-selected sets, the correlation was found to be close to 1. Social IQA achieved the lowest correlation value (0.982) between the two sets, while HellaSwag achieved the highest value (0.997).

Taken together, these results suggest that the surface irregularities in benchmarks likely do not explain the prior bias that we observed in our experiments. While a full causal analysis of prior bias is difficult to determine experimentally without a larger set of controls (comprising both carefully constructed benchmarks and a broader range of perturbation functions and confusion probes), at least one potential cause could be the \emph{hidden} patterns in the datasets used for fine-tuning. For example, it has been found that `annotation artifacts' \cite{gururangan2018annotation} can be introduced unintentionally by crowd workers constructing the benchmark (by devising hypotheses and candidate choices). Another hypothetical cause is that models may have picked up the bias during pre-training (e.g., due to increased frequency of some terms). A complete analysis of these hypothetical causes is left for future research.

\subsection{Additional Experiments Using UnifiedQA}
A major advance in the last few years has been the publication and release of universal `cross-format' multiple-choice NLI language models that have improved performance on various benchmarks without requiring fine-tuning per benchmark, as the RoBERTa model used in this paper required. Of these models, many of the latest ones designed for NLI are based on T5-11B (e.g., unifiedQA\cite{khashabi2020unifiedqa}). To verify the generalization of our findings for some of these newer models, we conducted preliminary experiments using pre-trained unifiedQA to replicate some of the findings described earlier using all four confusion probes.

Despite being trained already, the unifiedQA T5-11B model takes much longer to answer prompts due to the large size of its parameter space (in the billions). Therefore, in our preliminary experiments, we repeatedly sampled 50 instances three times from the \emph{dev.} set of each benchmark (for each intervention), and averaged the results to obtain stable performance estimates. The average original accuracy of unifiedQA on the four benchmarks is found to be 0.62. Considering a single model was used and no fine-tuning was done, this is a high performance compared to previous models tested with similar constraints. Comparative performance depends on the benchmark e.g., unifiedQA achieved similar performance (0.75) on PIQA, had a little performance decrease (0.77) on aNLI, but achieved lower performance on SocialIQA (0.52) and HellaSwag (0.42).

When applying the \emph{No-Question} probe, unifiedQA achieved a near-random performance on SocialIQA (0.38), just as RoBERTa had earlier. Encouragingly, unifiedQA had a more significant decrease in its performance on PIQA (the average `pseudo-accuracy' is 0.58), compared to RoBERTa (0.73). This may be a sign of progress in the NLI model's development, but the results would need to be replicated in the future for the full \emph{dev.} set, and not just the 50-instance sample (on which we averaged performance across three independent trials). On aNLI, unifiedQA obtained a similar `pseudo-accuracy' (0.59) compared to RoBERTa (0.64). On HellaSwag, unifiedQA's pseudo-accuracy decreased to 0.34, compared to RoBERTa's post-intervention pseudo-accuracy of 0.66. While the tendency is similar, unifiedQA achieved lower original performance on HellaSwag, and the percentage decline in unifiedQA's pseudo-accuracy compared to the original accuracy is much larger. This is again a promising sign but significant prior bias clearly still exists, although less than RoBERTa. 

When applying the \emph{Wrong-Question} probe, unifiedQA achieved a lower pseudo-accuracy compared to the \emph{No-Question} probe on Hellaswag and SocialIQA, but a higher pseudo-accuracy on PIQA and aNLI. The largest decrease was observed on HellaSwag, since the `pseudo-accuracy' changed from 0.34 (\emph{No-Question} probe) to 0.28 (\emph{Wrong-Question} probe).  On SocialIQA, there is a 0.04 decrease in unifiedQA's pseudo-accuracy compared to the \emph{No-Question} probe. On PIQA, however, the `pseudo-accuracy' \emph{increases} from 0.58 to 0.68 and on aNLI, the `pseudo-accuracy' had a 0.04 increase. Therefore, we again start to see benchmark-specific divergence in the behavior of a model. 


When applying the \emph{No-Right-Answer} probe, we verified that unifiedQA also preferred the \emph{non-substituted} options (which tend to be more contextually related to the prompt, even though they are incorrect choices), similar to what RoBERTa did, especially on PIQA. The probability of RoBERTa choosing the substituted option on PIQA is 0.37, but for unifiedQA, the probability drops to 0.26. On SocialIQA and HellaSwag, the two models behaved similarly for this probe. However, on aNLI, unifiedQA's `pseudo-accuracy' increases from 0.13 to 0.24, compared to ROBERTa. Once again, we observe benchmark-specific divergence. Besides, unifiedQA has a higher preference for semantically related (to the prompt), but still incorrect, answers compared to the fine-tuned RoBERTa.

Finally, when applying the \emph{Choice Paralysis} probe, the average accuracy of unifiedQA on each benchmark is near-random. This suggests, unlike the question-based interventions, that there is considerable room for improvement still since unifiedQA is obviously susceptible to choice paralysis. Following the 5-choice intervention, the accuracy of unifiedQA averaged across the four benchmarks is around 0.14, much lower than RoBERTa's near-original performance. Combined with the fact that unifiedQA exhibited disappointing performance on HellaSwag, the model may not be as capable of handling instances that have too many long-length choices, compared to instances and benchmarks that provide a relatively limited set of (2-3) short-length options.

\subsection{Additional Experiments Using XLNet}
XLNet\cite{yang2019xlnet} is an extension of the Transformer-XL model. It is pre-trained using an autoregressive method to learn bidirectional contexts. Under comparable experiment settings, XLNet was found to out-perform BERT on 20 tasks. Hence, we selected the pre-trained XLNet as another efficient model to conduct preliminary experiments using all four confusion probes. The version of XLNet we used is `xlnet-large-cased', provided by \cite{wolf-etal-2020-transformers}.  

Compared to unifiedQA, XLNet answers questions much faster; hence, we did not need to sample instances, but were able to replicate the experiments by just inputting all instances in the dev. set to XLNet. We found that, unlike unifiedQA, the non-fine-tuned XLNet only achieved random performance on the original dev. sets on all benchmarks (0.51 on aNLI, 0.52 on PIQA, 0.34 on Social IQA, and 0.24 on HellaSwag). After applying the No-Question, Wrong-Question, and No-Right-Answer probes, the `pseudo-accuracy' of XLNet on post-intervention instances remains random. The ideal result on the post-intervention instances is hence achieved at the cost of much lower performance on the pre-intervention instances, which is the inverse of what was observed for the equivalent fine-tuned RoBERTa model. When we implemented the 5-choice intervention, the average accuracy of XLNet across all benchmarks decreased to 0.17. While being slightly higher than unifiedQA, the performance is still lower than random.

\subsection{Additional Experiments Using Pre-trained RoBERTa}

To understand the dependencies between fine-tuning and the extent of prior bias and choice paralysis in RoBERTa, we also repeated these experiments using just the pre-trained RoBERTa model (i.e., that is not fine-tuned). We used a pre-trained RoBERTa model with a facility for multiple-choice classification, provided by \cite{wolf-etal-2020-transformers}. Similar to the pre-trained \emph{XLNet}, the pre-trained RoBERTa was also found to have near-random accuracy on the original dev. sets of all four benchmarks (0.51 on aNLI, 0.50 on PIQA, 0.34 on SocialIQA, and 0.28 on HellaSwag). Additionally, we found that, after applying the \emph{No-Question}, \emph{Wrong-Question} and \emph{No-Right-Answer} probes, the pre-trained RoBERTa model's `pseudo-accuracy' remained random. Interestingly, when we implemented the 5-choice intervention, the average accuracy of RoBERTa is 0.24, which is slightly higher than random (0.2). On select benchmarks, such as aNLI, the accuracy is as high as 0.3 (although still much lower than the ideal of 1.0). This result again supports the previous claim that `aNLI' is the easiest post-intervention benchmark for RoBERTa following application of the choice paralysis probe.

\section{Summary}

In this paper, we proposed to study the statistical prevalence of prior bias and choice paralysis in a popular language representation model based on BERT that has been extensively used in the commonsense reasoning community. Our methodology and experiments rely on publicly available benchmarks and multiple-choice NLI model, neither of which the authors had any role in developing or disseminating. We found evidence for both phenomena, although the extent of the phenomenon depends on the dataset and (in the case of choice paralysis) the experimental parameters used in the intervention. 

Further analysis, including an `irregularities analysis', suggests that these are complex phenomena not caused simply due to surface irregularities or artifacts in the prompts and choices. Interestingly, evidence of these phenomena is observed even in a more recent and independently pre-trained model, such as unifiedQA. Additional analyses reported in the \emph{Discussion} section, such as using the pre-trained RoBERTa model (without fine-tuning) and another transformer-based model called XLNet, show that the ideal bias-free performance (on perturbed instances) is achieved by these models, but at the cost of significantly lower performance on the original dev. set, which is the inverse of what was observed for the equivalent fine-tuned model. Namely, the benchmark-specific fine-tuned model achieved excellent performance on the unperturbed instances in the dev. set of the benchmark, but then exhibits prior bias on the perturbed instances.

In the future, we plan to extend the methodology to conduct detailed robustness studies of other linguistic phenomena and biases. Such studies provide important insights into the workings and biases of transformer-based models, even as they become more complex and widespread.

\newpage

%
%
%

%

\begin{acknowledgments}
This work was funded under the DARPA Machine Common Sense program.
\end{acknowledgments}

%

\bibliography{compling_style}

\end{document}